\documentclass[letterpaper, 10 pt, conference]{ieeeconf}
\IEEEoverridecommandlockouts
\usepackage{etextools}
\usepackage{amssymb}
\usepackage{float}
\usepackage{makeidx}
\usepackage{amsmath}
\usepackage{epsfig}
\usepackage{epsf}
\usepackage{psfrag}
\usepackage{verbatim}
\usepackage{color}
\usepackage{multirow}
\usepackage{tabularx}
\usepackage{booktabs}
\usepackage[tight,footnotesize]{subfigure}
\usepackage{array}
\usepackage{soul}
\usepackage{footnote}
\usepackage{cite}
\usepackage{dblfloatfix}

\usepackage{graphicx}
\graphicspath{{Figures}}
\DeclareGraphicsExtensions{.pdf,.png}
\usepackage{color}
\usepackage{multirow}

\usepackage{hyperref}
\hypersetup{
    colorlinks=true,
    citecolor=blue
}
\usepackage{bm}

\usepackage{bigstrut}

\title{\LARGE \bf
Robots that Take Advantage of Human Trust
}

\author{Dylan P. Losey and Dorsa Sadigh
\thanks{The authors are with the Stanford Intelligent and Interactive Autonomous Systems Group (ILIAD), Department of Computer Science, Stanford University, Stanford, CA 94305.
	{(e-mail: dlosey@stanford.edu)}}%
}

\begin{document}

\maketitle

\begin{abstract}
    Humans often assume that robots are rational. We believe robots take optimal actions given their objective; hence, when we are uncertain about what the robot's objective is, we interpret the robot's actions as optimal with respect to our estimate of its objective. This approach makes sense when robots straightforwardly optimize their objective, and enables humans to learn what the robot is trying to achieve. However, our insight is that---when robots are aware that humans learn by \textit{trusting} that the robot actions are rational---intelligent robots do not act as the human expects; instead, they take advantage of the human's trust, and exploit this trust to more efficiently optimize their own objective. In this paper, we formally model instances of human-robot interaction (HRI) where the human does not know the robot's objective using a two-player game. We formulate different ways in which the robot can model the uncertain human, and compare solutions of this game when the robot has conservative, optimistic, rational, and trusting human models. In an offline linear-quadratic case study and a real-time user study, we show that trusting human models can naturally lead to communicative robot behavior, which influences end-users and increases their involvement.
\end{abstract}

\section{Introduction}
\label{sec:intro}

When humans interact with robots, we often assume that the robot is rational \cite{baker2009}. Imagine a human helping a robot to set the table. The human end-user has told the robot to bring two plates from the kitchen; but, when the robot returns, it is only carrying one. Based on this action---and our assumption that robots are rational---the end-user learns about the robot's objective. For instance, here the human infers that the robot assigns a high cost to carrying multiple plates at once.

A level of human \emph{trust} is implicit within this interpretation. If we assume that the robot takes actions purely to optimize its objective, we trust that these actions are \emph{not meant} to influence the user. In our example, a trusting human does not realize that the robot may actually be manipulating them into carrying the second plate. This robot is capable of bringing two plates, but misleads the human to believe that it can only carry one in order to reduce its overall effort. Hence, trusting the robot leads to incorrect human learning: the human does not learn the robot's true objective, but instead the objective the robot has chosen to convey. This behavior is useful for assistive robots, where humans can become overly reliant on the robot's assistance \cite{blank2014}, and robots need to make intelligent decisions that motivate user participation.

Within this work, we study HRI where the human does not know the robot's objective. We recognize that:
\vspace{-0.5em}\begin{center}
    \emph{If the robot is aware of the human's trust, an optimal robot should select its actions so as to} communicate \emph{with the human and} influence \emph{their understanding.}
\vspace{-0.5em}\end{center}
Our insight results in autonomous agents that intentionally communicate with nearby humans as a part of their optimal behavior. In practice, however, we find that it is not always in the robot's best interests to communicate its true objective; by contrast, \emph{misleading actions naturally emerge} when robots model humans as trusting agents. This indicates that---if the robot has a different objective than the human---user trust becomes something that the robot can exploit.

Overall, we make the following contributions:

\smallskip

\noindent\textbf{Formulating Human Models.} We formulate four ways that robots can model uncertain humans within a two-player HRI game. The robot can behave conservatively (as if the human already knew their objective), optimistically (as if they can teach the human any objective), model the human as rational, or treat the human as a trusting agent. Trusting is a simplification of the rational model: the human learns while assuming that the robot purely optimizes its objective.

\smallskip

\noindent\textbf{Comparison Between Models.} We show that modeling the human as trusting is not the same as acting optimistically, but can result in robots that take advantage of the human.

\smallskip

\noindent\textbf{Experiments.} To understand the relationship between trusting human models and communicative or influential robot behavior, we perform an offline linear-quadratic case study and a real-time user study. Unlike the conservative and optimistic human models, communicative robot behavior naturally emerges when the robot models the human as trusting, causing the human participants to adapt their behavior and increase their overall involvement.

\smallskip

\noindent This work is a contribution towards developing communicative robots, and explores the interplay between how humans model robots and how optimal robots exploit these models.

\begin{figure*}[t]

	\begin{center}
		\includegraphics[width=1.8\columnwidth]{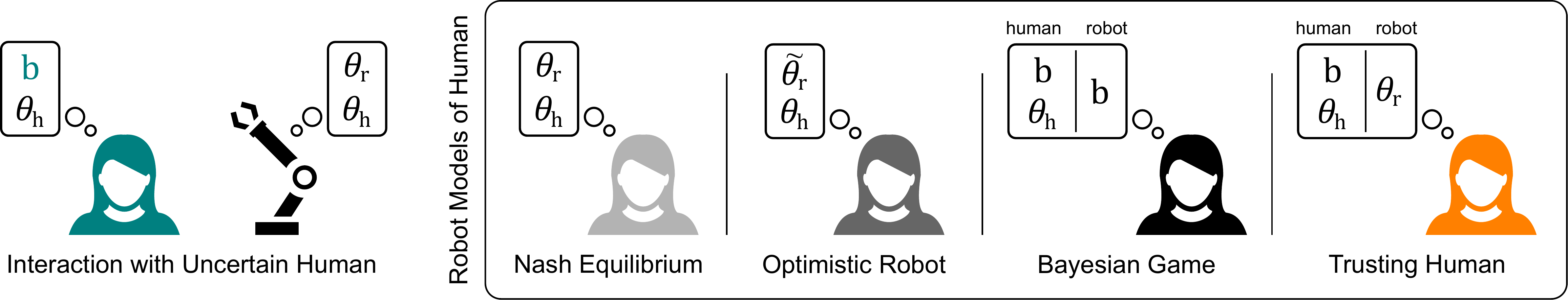}

		\caption{Human-robot interaction where the human does not know the robot's objective, $\theta_r$. The robot chooses its behavior based on its model of the human. A conservative robot (Nash) acts as if the human knew $\theta_r$, while a optimistic robot (Optimistic) thinks that it can teach the human an arbitrary objective $\tilde{\theta}_r$. Rather than acting conservatively or optimistically, the robot can treat the human as a rational agent (Bayesian), where the human recognizes that the robot is aware of their uncertainty $b$. Because real humans are not completely rational, we propose a simpler alternative (Trusting): here the human learns about $\theta_r$ from the robot while assuming that the robot behaves conservatively.}

		\label{fig:front}
	\end{center}

\vspace{-2em}

\end{figure*}

\section{Related Work}

Previous research has studied trust during HRI, along with the robotic and environmental factors that affect the human's trust \cite{hancock2011}. Autonomous agents can model and infer the human's trust based on past interactions \cite{xu2015}, and increase the human's trust by making the robot's actions transparent to explain its decision making process \cite{wang2016}. Most relevant here is work by Chen \textit{et al}. \cite{chen2018}, who treat the human's trust as a latent variable, and integrate trust into the robot's decision making. These previous works \cite{hancock2011, xu2015, wang2016, chen2018} consider cases where the human and collaborative robot share a common objective and the robot should maximize trust; by contrast, we are interested in robots that leverage trust to influence users.

Other research focuses on robots that utilize their actions to implicitly communicate with nearby humans. Communicative robot actions facilitate human understanding during the task \cite{hellstrom2018} so that humans can anticipate what the robot will do in the future. In order to identify communicative actions, we need a cognitive model of how the human learns: the human might respond to the robot's actions \cite{sadigh2016}, infer the robot's objective \cite{huang2017}, or follow a personalized learning strategy \cite{losey2019}. Our research builds upon these prior works \cite{hellstrom2018,sadigh2016,huang2017,losey2019,sadigh2018planning} by identifying communicative actions for a trusting human model. Unlike research that explicitly generates deceptive actions \cite{dragan2014}, here the robot naturally selects misleading communication. This is similar to learning with opponent-learning awareness \cite{foerster2018}: here only the robot is aware.

We can also frame our work as a contribution to perspective taking, where the robot reasons about the human's perspective while make decisions \cite{berlin2006, pandey2013, trafton2005}. Prior research in perspective taking considers which objects are accessible to the human (i.e., what can the human see or reach), and how the robot can reduce human confusion or communicate its intention through proactive, contextual behavior. In our work, the robot assumes the human has a trusting perspective (or theory of mind), and solves for optimal responses given this human perspective.

\section{Problem Formulation}

Consider a human interacting with a robot. The human and robot each have their own objective, and take actions in order to achieve that objective. We are interested in settings where the robot \emph{knows} both the robot and human objectives, but the human \emph{does not know} the robot's objective. While many prior works (e.g., imitation learning) focus on expert humans---who already have access to all the necessary information, and must teach that information to the robot learner---here we explore the converse situation.

In this section, we first model human-robot interaction as a two-player game. Next, we formulate four different strategies for selecting robot actions, where each strategy is optimal given the corresponding human model (see Fig.~\ref{fig:front}). Finally, we discuss the plate carrying example mentioned in Sec.~\ref{sec:intro} to compare these alternative formulations.

\subsection{Human Robot Interaction as a Two-Player Game} \label{sec:game}

We model human-robot interaction (HRI) as a two-player Markov game, where both agents affect the state of a dynamical system. Let the system have state $x \in \mathcal{X}$, let the robot take action $u_r \in \mathcal{U}_R$, and let the human take action $u_h \in \mathcal{U}_H$. The state transitions according to: $x^{t+1} = f(x^t, u_r^t, u_h^t)$, where $f$ captures the system dynamics. This game starts at $t=0$, and ends after $H \in \mathbb{Z}^+$ timesteps when $t=H$.

\smallskip

\noindent\textbf{Rewards.} The robot has a reward function: $r_r(x, u_r, u_h, \theta_r)$, and the human has a reward function: $r_h(x, u_r, u_h, \theta_h)$. These reward functions are parameterized by the constants $\theta_r$ and $\theta_h$, which capture the robot and human objectives, respectively. Notice that we do not assume that $\theta_h = \theta_r\,$; hence, the human and robot may have different objectives.

\smallskip

\noindent\textbf{Game.} During each timestep the human and robot observe the current state $x^t$ and take simultaneous actions $u_h^t$ and $u_r^t$. The robot receives reward $r_r(x^t, u_r^t, u_h^t, \theta_r)$ and the human receives reward $r_h(x^t, u_r^t, u_h^t, \theta_h)$. Both agents are attempting to maximize their own sum of rewards over the finite horizon $H$. Denote the sequence of robot actions as $\bm{u_r} = \{u_r^0, \ldots, u_r^H\}$ and the sequence of human actions as $\bm{u_h} = \{u_h^0, \ldots, u_h^H\}$. The sum of rewards for the robot and human, respectively, are:
\begin{equation}
    R_r(x^0, \bm{u_r}, \bm{u_h}, \theta_r) = \sum_{t=0}^H r_r(x^t, u_r^t, u_h^t, \theta_r)
    \label{eq:Rr}
\end{equation}
\begin{equation}
    R_h(x^0, \bm{u_r}, \bm{u_h}, \theta_h) = \sum_{t=0}^H r_h(x^t, u_r^t, u_h^t, \theta_h)
    \label{eq:Rh}
\end{equation}

\smallskip

\noindent \textbf{Beliefs over reward functions.} The robot knows both $\theta_r$ and $\theta_h$, while the human only has access to its own reward parameters $\theta_h$. Let $\mathcal{D}$ be the probability distribution from which the parameter $\theta_r$ is sampled; we assume that the human is aware of this distribution $\mathcal{D}$, and updates their belief over the robot's objective based on the robot's actions. More specifically, the human has a belief: $b^{t+1}(\theta_r) = P(\theta_r ~ | ~ x^{0:t}, u_r^{0:t}, u_h^{0:t}, \theta_h, \mathcal{D})$.

\subsection{Formulating Robot's Responses to Humans}

Given our description of HRI as a two-player game, we want to plan for the robot's optimal behavior. Importantly, the robot's optimal behavior depends on \emph{how the robot models the human}. Recall our plate carrying example: if the robot models the human as a rational agent (where the robot cannot mislead the user) the robot should bring two plates, but if the robot models the human as a trusting agent, it should only bring one. Below we formulate optimal robot behavior in response to \emph{four different} human models within the context of a two-player game.

\medskip

\noindent \textbf{1) Nash Equilibrium.} This robot acts as if the human also knows $\theta_r$, so that both the human and robot objectives are common knowledge. When each agent has all the information, their optimal behavior can be computed by finding the Nash equilibrium of this game. Throughout this work, we assume that a pure-strategy Nash equilibrium exists, and---if there are multiple equilibria---both agents coordinate to follow the same one. More formally, $(\bm{u_r}^{*,n}, \bm{u_h}^{*,n})$ is a strict pure-strategy Nash equilibrium over the finite horizon $H$ if:
\begin{equation} \label{eq:fn}
\begin{gathered}
    \bm{u_r}^{*,n} = \text{arg}\max_{\bm{u_r}} ~ R_r\big(x^0, \bm{u_r}, \bm{u_h}^{*,n}(\theta_r, \theta_h), \theta_r \big)
\\
    \bm{u_h}^{*,n} = \text{arg}\max_{\bm{u_h}} ~ R_h\big(x^0, \bm{u_r}^{*,n}(\theta_r, \theta_h), \bm{u_h}, \theta_h \big)
    \end{gathered}
\end{equation}
We emphasize that the human and robot policies $\bm{u_h}^{*,n}(\theta_r, \theta_h)$ and $\bm{u_r}^{*,n}(\theta_r, \theta_h)$ can be specified as a function of the reward parameters $\theta_r$ and $\theta_h$ since these policies are optimizers of the reward functions (\ref{eq:Rr}) and (\ref{eq:Rh}). 

A robot should follow $\bm{u_r}^{*,n}$ when the human knows $\theta_r$. But, in our setting, the human does not initially know $\theta_r\,$; hence, we can think of $\bm{u_r}^{*,n}$ as a \emph{conservative} robot, which never tries to exploit the human's uncertainty.

\medskip

\noindent \textbf{2) Optimistic Robot.} At the other end of the spectrum, the robot can behave \emph{optimistically}, and assume that---because the human does not know the robot's objective---the robot can teach the human an arbitrary objective $\tilde{\theta}_r\,$:
\begin{equation} \label{eq:fs}
\begin{gathered}
    \bm{u_r}^{*,s} = \text{arg}\max_{\bm{u_r}} ~ R_r\big(x^0, \bm{u_r}, \bm{u_h}^{*,s}(\tilde{\theta}_r, \theta_h), \theta_r \big)
\\
    \bm{u_h}^{*,s} = \text{arg}\max_{\bm{u_h}} ~ R_h\big(x^0, \bm{u_r}^{*,n}(\tilde{\theta}_r, \theta_h), \bm{u_h}, \theta_h \big)
\end{gathered}
\end{equation}
where the robot picks $\tilde{\theta}_r$. Comparing (\ref{eq:fs}) to (\ref{eq:fn}), here the human responds as if the robot's objective is $\tilde{\theta}_r$ rather than the true objective $\theta_r$.

\medskip

\noindent \textbf{3) Bayesian Game.} Instead of acting conservatively or optimistically, the robot can model the human as a \emph{rational} agent, where this human solves the HRI game. Given that the human has a belief $b$ over the robot's objective, we formalize rational human behavior as a Bayesian equilibrium \cite{osborne1994}:
\begin{equation} \label{eq:fb}
\begin{gathered}
    \bm{u_r}^{*,b} = \text{arg}\max_{\bm{u_r}} ~ R_r\big(x^0, \bm{u_r}, \bm{u_h}^{*,b}(b(\theta_r), \theta_h), \theta_r \big)
\\
    \bm{u_h}^{*,b} = \text{arg}\max_{\bm{u_h}} ~ \mathbb{E}_{\theta \sim b} \Big[R_h\big(x^0, \bm{u_r}^{*,b}(\theta, \theta_h), \bm{u_h}, \theta_h \big)\Big]
\end{gathered}
\end{equation}
$b$ updates based on the robot's actions using Bayesian inference, i.e., 
 $b^{t+1}(\theta_r) = b^t(\theta_r) \cdot P(u_r^t ~| ~ x^{0:t}, u_r^{0:t}, u_h^{0:t}, \theta_h, \theta_r)$.

Notice that, within a Bayesian game, the human realizes that (a) the robot knows about its belief $b$, and (b) the robot takes actions while considering this belief.

\medskip

\noindent \textbf{4) Trusting Human.} In reality, humans are not completely rational \cite{hedden2002,simon1997}. Instead of modeling the human as rational, the robot can leverage a simplified model where the human is \emph{trusting}. This trusting human assumes that---even though they do not know the robot's reward function---\emph{the robot will act conservatively, as if the human knows} $\theta_r$:
\begin{equation} \label{eq:ft}
\begin{gathered}
    \bm{u_r}^{*,\tau} = \text{arg}\max_{\bm{u_r}} ~ R_r\big(x^0, \bm{u_r}, \bm{u_h}^{*,\tau}(b(\theta_r), \theta_h), \theta_r \big)
\\
    \bm{u_h}^{*,\tau} = \text{arg}\max_{\bm{u_h}} ~ \mathbb{E}_{\theta \sim b} \Big[R_h\big(x^0, \bm{u_r}^{*,n}(\theta, \theta_h), \bm{u_h}, \theta_h \big)\Big]
\end{gathered}
\end{equation}
Comparing (\ref{eq:ft}) to (\ref{eq:fb}), a trusting human assumes that the robot follows $\bm{u_r}^{*,n}$, while a rational human understands that the robot reasons over their belief $b$. 

Let us return to our plate carrying example. A rational human, $\bm{u_h}^{*,b}$, recognizes that the robot may be trying to mislead them, and is therefore always reluctant to carry a plate. By contrast, a trusting human, $\bm{u_h}^{*,\tau}$, assumes that the conservative robot will never try to take advantage of their uncertainty: if the robot only brings a single plate, this indicates to the trusting human that robot cannot carry more, and so the user carries a second plate to assist the robot. We present a more formal description of this example below.

\subsection{Plate Carrying Example}

A human and a robot are carrying plates to set the table. Both agents want more plates to be placed upon the table, but prefer for the other agent to bring these plates. The system state in this problem is $x$, where $x \in \mathbb{Z}^+$ is the total number of plates on the table. During every timestep, the robot and human can each bring up to two new plates: $\mathcal{U}_R = \mathcal{U}_H = \{0, 1, 2\}$, and hence the system dynamics are: $x^{t+1} = x^t + u_r^t + u_h^t$. We let $\theta_r, \theta_h > 0$, and define the robot and human rewards so that $\theta_r$ and $\theta_h$ individually penalize the robot and human for carrying plates. More specifically, for each timestep $t < H$: $r_r(x, u_r, u_h, \theta_r) = - \theta_r u_r^2$ and $r_h(x, u_r, u_h, \theta_h) = - \theta_h u_h^2$. At the final timestep $t=H$ each agent is rewarded for the total number of plates set on the table: $r^H(x) = x^H$. For simplicity, we constrain this example problem so that the human and robot cannot carry plates at the same time. To do this, we added a penalty $-\alpha \cdot u_r u_h$ to both $r_r$ and $r_h$, where $\alpha \rightarrow \infty$ makes it prohibitively expensive for the agents to carry plates simultaneously.

In the next section we list how the robot should optimally carry plates for different human models. 

\section{Planning for Interaction \\ with Uncertain Humans}

\begin{table*}[t]
	\caption{Example of robot interactions with uncertain humans under different formulations. The robot and human are carrying plates: one interesting case occurs when $\theta_r = 0.2$ and $\theta_h = 0.25$. Here a robot interacting with a trusting human only brings a single plate at the first timestep, influencing the human to carry plates during the second timestep.}
	\label{table}
	\begin{center}
		\begin{tabular}{{c}|{c}{c}{c}{c}{c}{c}}
Formulation & \textbf{Nash} & \textbf{Optimistic} & \textbf{Bayesian} & \textbf{Trusting} & \textbf{Bayesian} & \textbf{Trusting}\\ 
Distribution $\mathcal{D}$ & $-$ & $-$ & $\mathcal{U}(0, 1.5)$ & $\mathcal{U}(0, 1.5)$ & $Boltzmann(a = 0.1)$ & $Boltzmann(a = 0.1)$ \bigstrut \\ \hline
Robot actions at $t=0,1$ & $2, 2$ & $0, 0$ & $1, 0$ & $1, 0$ & $2, 2$ & $1, 0$  \bigstrut[t] \\
Human actions at $t=0,1$ & $0, 0$ & $2, 2$ & $0, 2$ & $0, 2$ & $0, 0$ & $0, 2$ \\
Mean human estimate of $\theta_r$ & $0.2$ & $> 1$ & $4/7$ & $2/3$ & $0.22$ & $0.43$ \\
$R_r$, $R_h$ & $2.4$, $4$ & $4$, $2$ & $2.8$, $2$ & $2.8$, $2$ & $2.4$, $4$ & $2.8$, $2$
		\end{tabular}
	\end{center}

	\vspace{-2em}	

\end{table*}

We have introduced four formulations for identifying robot behavior when the human does not know the robot's objective, where each formulation is optimal given a different human model: conservative, optimistic, rational, and trusting. Next, we focus on how robots that model humans as trusting agents \emph{compare} to these alternative formulations. We first demonstrate that we can express HRI with a trusting human as a Markov decision process, and, accordingly, we can leverage existing techniques to plan for interactions with trusting humans that do not know $\theta_r$. We then compare the trusting human model to Bayesian games using our plate carrying example, and show that---while responding to trusting humans is not the same as acting optimistically---an intelligent robot \emph{takes advantage} of the human's trust.

\subsection{Trusting Humans as a Markov Decision Process}

Here, we revisit the two-player game outlined in Section~\ref{sec:game}, and use the trusting human model (\ref{eq:ft}) to reformulate this game as an instance of a single-agent Markov decision process (MDP). Expressing the trusting human model as an MDP enables us to leverage existing methods to solve for $\bm{u_r}^{*,\tau}$. Our insight is recognizing that---instead of the state $x$---this MDP has an augmented state $s$ that captures the necessary information to model the trusting human.

\smallskip

\noindent \textbf{MDP.} Interacting with trusting humans that do not know the robot's objective is a finite-horizon MDP defined by the tuple $\mathcal{M} = \langle \mathcal{S}, \mathcal{U}_R, T, \theta_r, H \rangle$ \cite{puterman2014}. We introduced some of these terms in Section~\ref{sec:game}: $\mathcal{U}_R$ is the robot's action space, $\theta_r$ is the robot's objective, and $H$ is the finite planning horizon.

\smallskip

\noindent \textbf{State.} Define the augmented state $s \in \mathcal{S}$ to be $s = (x, \theta_h, b)$, where $x$ is the system state, $\theta_h$ is the human's objective, and $b$ is the human's belief over the robot's objective. 

\smallskip

\noindent \textbf{Prediction.} Given $s^t$ the robot can predict the current human action $u_h^t$ under the trusting human model. First, using $\theta_h$, the robot solves (\ref{eq:fn}) to obtain $\bm{u_r}^{*,n}(\theta, \theta_h)$ for all $\theta_r \in \Theta_R$. Next, using $b^t$, $x^t$, $\bm{u_r}^{*,n}(\theta, \theta_h)$, and $\theta_h$, the robot obtains $\bm{u_h}^{*,\tau}$ with (\ref{eq:ft}). $\bm{u_h}^{*,\tau}$ is a sequence; the $t$-th element of $\bm{u_h}^{*,\tau}$ is $u_h^t$. Repeating this process at every $s$ yields the human policy.

\smallskip

\noindent \textbf{Transition.} The transition function $T(s^{t+1} ~ | ~ s^t, u_r^t)$ captures how $s$ evolves over time. We previously defined the system dynamics as $x^{t+1} = f(x^t, u_r^t, u_h^t)$, $\theta_h$ is a constant, and $b$ updates based on how the human learns (e.g., Bayesian inference or gradient descent).

\smallskip

\noindent\textbf{Underactuated Dynamics of Interaction.} When uncertain humans are trusting agents, the interaction between the human and robot becomes an \emph{underactuated} system. This comes from our MDP reformulation: the trusting human policy is a function of $s$, and the robot employs its actions $u_r$ to control how $s$ transitions. Consider our plate carrying example: if the robot brings two plates, the human will carry no plates, but if the robot brings a single plate, the human carries the second plate. Hence, the robot uses its own actions to indirectly control (i.e., influence) the human's behavior.

\subsection{Comparing Trusting Humans to Other Human Models}

Now that we know how to identify optimal behavior when interacting with trusting humans, we explore how this behavior compares to interacting under other human models. Using our plate carrying example, we compute how uncertain humans and conservative (Nash) or optimistic robots interact, as well as robots that model the human as completely rational (Bayesian). By comparing examples of this behavior to our trusting model, we demonstrate that there are situations where robots take advantage of the human's trust.

\smallskip

\noindent\textbf{Computing Solutions.} We list the robot and human behavior for one interesting case of our plate carrying example in Table~\ref{table}. When $\theta_h, \theta_r < 1/3$, we conservatively assume that the agents follow the Nash equilibrium (\ref{eq:fn}) where the robot carries the plates. For Optimistic (\ref{eq:fs}), the robot chooses $\tilde{\theta}_r > 1$, so that it always makes sense for the human to carry plates.

We list the Bayesian (\ref{eq:fb}) and Trusting (\ref{eq:ft}) models under two different distributions $\mathcal{D}$. Intuitively, when $\theta_r$ is sampled from a uniform distribution the rational human is very uncertain about the robot's objective, and so the robot can induce the human to carry plates. On the other hand, when $\theta_r$ is sampled from a Boltzmann distribution, the human is confident that the robot's objective is low, and thus the robot becomes unable to mislead this rational human. More specifically, a rational human recognizes that the Bayesian robot plays $u_r = 1$ for $\theta_r \in [1/7,1]$ with a uniform prior, and learns a mean estimate $4/7$. A trusting human thinks the robot only plays $u_r = 1$ for $\theta_r \in [1/3,1]$, and so their mean estimate becomes $2/3$ with a uniform prior. These computations are similar for the Boltzmann $\mathcal{D}$, but, because the human is confident that $\theta_r < 1/3$, playing $u_r = 1$ is not sufficient to convince a rational human that $\theta_r > 1/3$.

\smallskip

\noindent\textbf{Analysis.} The results from our the plate carrying example provide a useful comparison between the trusting human model and our other formulations.

First, we observe that interacting with a trusting human is not necessarily the same as behaving optimistically. Consider the example in Table~\ref{table}: while the optimistic robot plans for the best possible case---where the human carries all the plates---the robot interacting with a trusting end-user recognizes that the human's actions depend on their belief. This robot reasons about how its behavior influences the human's understanding of the robot's objective: within the example, the robot initially brings a single plate to convince the human to carry additional plates at the next timestep.

Second, there are situations where robots interacting with trusting humans take advantage of that human's trust. Consider the Boltzmann distribution case in Table~\ref{table}: when the human trusts the robot, the robot intentionally misleads that human into carrying additional plates, but, when the human is rational (Bayesian), the robot carries all the plates. In this case, being trusting leads to less reward for the human than being rational: $R_h(x^0, \bm{u_r}^{*,\tau}, \bm{u_r}^{*,\tau}, \theta_h) < R_h(x^0, \bm{u_r}^{*,b}, \bm{u_r}^{*,b}, \theta_h)$.

\smallskip

\noindent\textbf{Summary.} When comparing robots planning under a trusting human model to Bayesian games, we found that interacting with trusting humans is not the same as a robot behaving optimistically, but it can lead to cases where trusting humans receive less reward than rational humans.

\section{Case Study}

\begin{figure}[t]

\vspace{0.5em}

	\begin{center}
		\includegraphics[width=0.8\columnwidth]{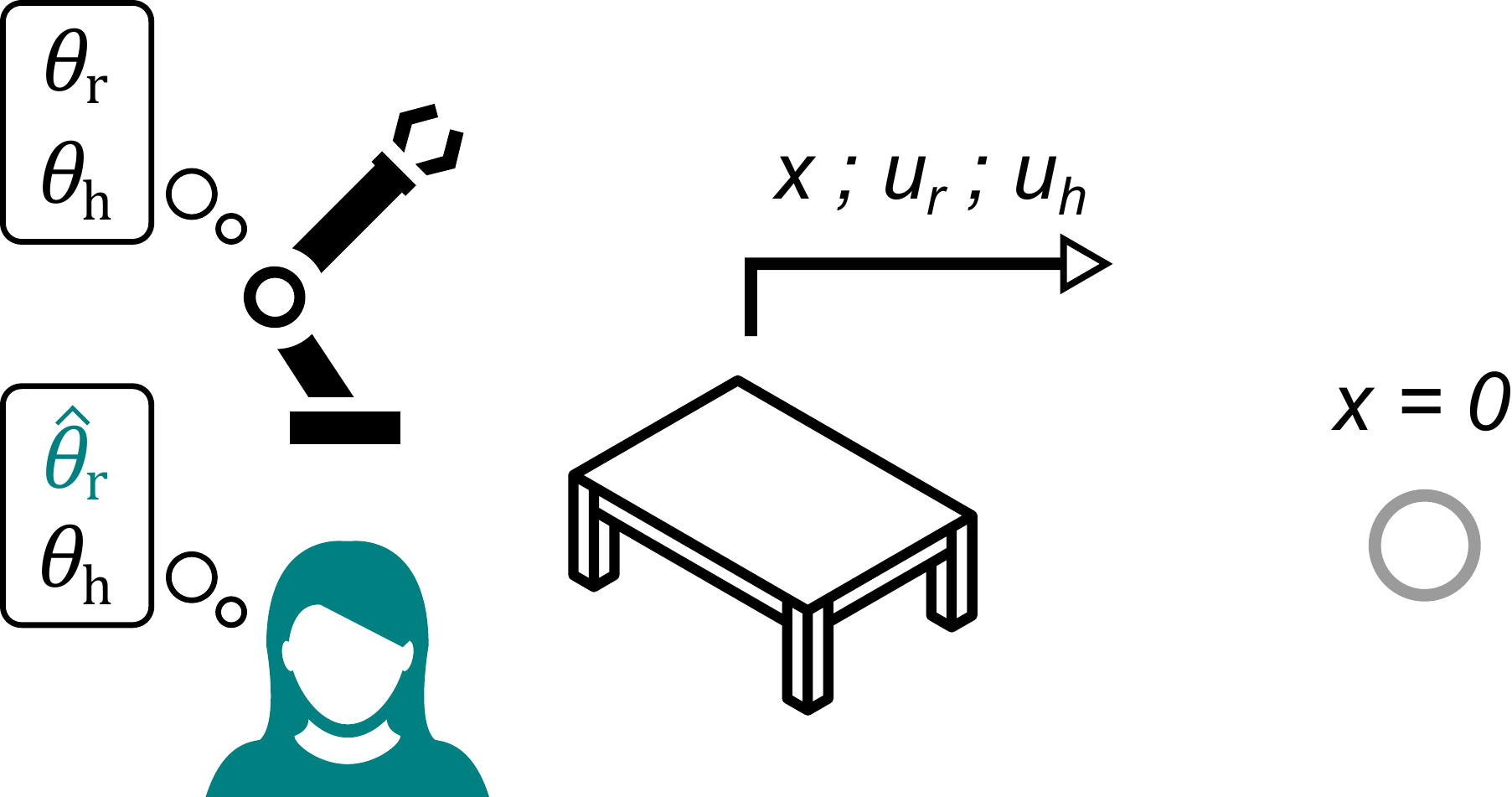}

		\caption{Setup for our case study. The human and robot are pushing a table with linear dynamics to the goal state ($x = 0$). Both agents want to minimize a quadratic cost function that considers (a) error from the goal and (b) their own effort when pushing the table. The human has a point estimate $\hat{\theta}_r$ of $\theta_r$, which the human updates using gradient descent with learning rate $\eta$.}

		\label{fig:simworld}
	\end{center}

\vspace{-1.5em}

\end{figure}

\begin{figure*}[t]

\vspace{0.5em}

	\begin{center}
		\includegraphics[width=1.95\columnwidth]{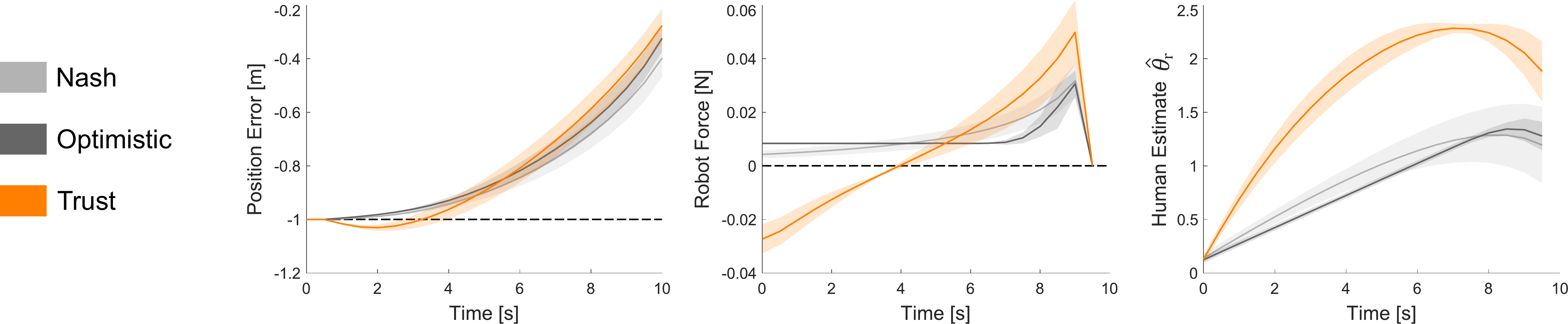}

		\caption{Example interaction with a trusting human that learns from robot actions (learning rate $\eta = 5$). Left: the position of the table with respect to the goal. Middle: the robot's applied force. Right: the human's estimate $\hat{\theta}_r$ of the robot's true objective $\theta_r$. Shaded regions show the standard error of the mean (SEM) across trials. The \emph{Trust} robot is aware of the human's trust, and intentionally moves the table away from the goal at the start of the task in order to convince the human to complete the rest of the task autonomously. Although \emph{Trust} applies more total effort than \emph{Nash} and \emph{Optimistic}, it ultimately obtains a higher reward since it quickly convinces the human to take over, and thus completes the task with the least state error. Notice that the human learned an increased value for $\hat{\theta}_r$ when the robot applied small forces towards the goal (\emph{Nash} and \emph{Optimistic}), but pulling the table away from the goal (\emph{Trust}) resulted in the largest human estimate $\hat{\theta}_r$, better misleading the human into believing that the robot was uninterested in the goal.}

		\label{fig:result}
	\end{center}

\vspace{-0.5em}

\end{figure*}

\begin{figure*}[t]

	\begin{center}
		\includegraphics[width=1.8\columnwidth]{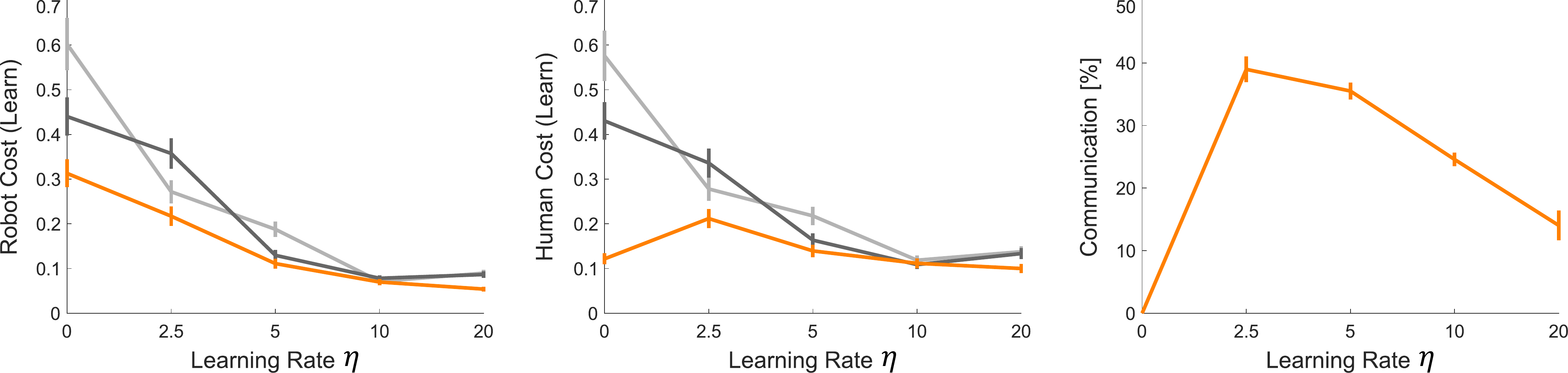}

		\caption{Objective performance when interacting with a trusting human during a linear-quadratic game. Left: the cost incurred by the robot. Middle: the cost incurred by the human. Right: the percentage of robot actions that are communicative. Error bars report the standard error of the mean (SEM). \emph{Trust} leads to less robot cost than either \emph{Nash} or \emph{Optimistic}, and also reduces the cost for the human agent. Only the \emph{Trust} robot takes communicative actions, i.e., actions that move the table away from the goal. Although these actions are locally sub-optimal, they convey the robot's future behavior to the human, and improve the overall HRI performance (in terms of cost) for both the robot and human.}

		\label{fig:learn}
	\end{center}

\vspace{-2.0em}

\end{figure*}

To understand how modeling the human as a trusting agent is related to \emph{communicative} robot behavior, we performed a simulated user study. This study focuses on a \emph{linear-quadratic} (LQ) problem, where the system dynamics are linear and the human and robot reward functions are quadratic: our problem setup is shown in Fig.~\ref{fig:simworld}. Here the robot knows both the human and robot objectives, but the human only has a point estimate of the robot's objective. We first considered humans that learn about the robot's objective from the robot's behavior, and demonstrate how our approach leads to robots that purposely communicate with and mislead the trusting human. Next, we added users that did not follow our trusting human model, and measured performance during interactions when the trusting human model was inaccurate.

\subsection{Simulation Setup}

During each simulation, the human and robot pushed a one degree-of-freedom mass-damper table towards a goal state. Let $x = [p, v]^\top \in \mathbb{R}^2$ be the position $p$ and velocity $v$ relative to the goal state, let $m$ be the mass of the table, and let $b$ be a viscous friction constant. The linear system dynamics are:
\begin{equation} \label{eq:E1}
    \begin{gathered}
    x^{t+1} = Ax^t + B(u_r^t, u_h^t) \\[1.5ex] 
    A = \begin{bmatrix} 1 & \Delta \\ 0 & 1 - {b\Delta}/{m} \end{bmatrix}, \quad B = \begin{bmatrix} 0 \\ \Delta/m \end{bmatrix}
    \end{gathered}
\end{equation}
where $\Delta$ is the period between $t$ and $t+1$. The agents wanted to minimize the distance of table from the goal at timestep $H$, while also minimizing their own in-task effort:
\begin{equation} \label{eq:E2}
    R_r(x^0, \bm{u_r}, \bm{u_h}, \theta_r) = -\|x^H\|^2 - \theta_r \cdot \Delta \bm{u_r}^\top\bm{u_r}
\end{equation}
\begin{equation} \label{eq:E3}
    R_h(x^0, \bm{u_r}, \bm{u_h}, \theta_h) = -\|x^H\|^2 - \theta_h \cdot \Delta \bm{u_h}^\top\bm{u_h}
\end{equation}
Here $\theta_r, \theta_h \in \mathbb{R}$ are constants that determine the trade-off between error and effort. The robot knew $\theta_r$ and $\theta_h$, while the human only knew $\theta_h$ and had a point estimate $\hat{\theta}_r$ of $\theta_r$. Similar to prior work~\cite{liu2017}, we assumed that the human updated their point estimate using gradient descent:
\begin{equation} \label{eq:gd}
    \hat{\theta}_r^{t+1} = \hat{\theta}_r^t + \eta \cdot \nabla_{\theta}\big| \bm{u_r}^{*,n}(\theta,\theta_h) - u_r^t\big|
\end{equation}
with learning rate $\eta > 0$. Thus, the human learned based on the difference between the predicted robot behavior, $\bm{u_r}^{*,n}$, and the actual robot action, $u_r$. When both agents know $\theta_r$ and $\theta_h$, this HRI problem is an instance of a deterministic linear-quadratic game \cite{engwerda2007}. Control strategies that achieve a Nash equilibrium are of the form $u_r^t = -K_r^t x^t$ and $u_h^t = -K_h^t x^t$, where $K_r, K_h \in \mathbb{R}^{1 \times 2}$ are feedback gain matrices \cite{engwerda2007, li2019}. We note that---within this deterministic LQ game---a pure Nash equilibrium exists and is unique.

\subsection{Independent Variables}

We compared three formulations for selecting robot behavior: conservatively following the Nash equilibrium (\emph{Nash}), optimistically solving the two-player game (\emph{Optimistic}), or computing optimal behavior for a trusting human (\emph{Trust}). For \emph{Nash} the robot follows (\ref{eq:fn}), for \emph{Optimistic} the robot solves (\ref{eq:fs}), and for \emph{Trust} the robot leverages (\ref{eq:ft}). Each of these robot strategies makes a different assumption about the human user: \emph{Nash} acts as if the human knew $\theta_r$, \emph{Optimistic} assumes the human believes the robot objective is $\tilde{\theta}_r$, and \emph{Trust} simplifies the rational human model.

To assess the performance of these robotic approaches with different types of users, we also varied the human's interaction strategy. There were three types of simulated users:
\begin{itemize}
    \item \textbf{Learn,} a trusting end-user that updates its point estimate $\hat{\theta}_r$ using (\ref{eq:gd}). Within this category we simulated humans that learn at different rates: $\eta = \{0, 2.5, 5, 10, 20\}$.
    \item \textbf{Fixed,} an end-user who assumes $\hat{\theta}_r^{t+1} = \hat{\theta}_r^{t}$, and always follows the Nash equilibrium for their initial estimate of $\theta_r$. We can think of this as an end-user who never adapts to the robot during HRI. Fixed is the same as Learn when $\eta = 0$, but we use Fixed to test situations where the robot incorrectly believes $\eta > 0$ (e.g., $\eta = \{5, 10, 20\}$).
    \item \textbf{Predict,} an end-user who predicts that the next $N$ robot actions will be the same as $u_r^t$, and responds accordingly.
\end{itemize}
\noindent Our simulations explored how \emph{Nash}, \emph{Optimistic}, and \emph{Trust} interact with these different types of simulated humans.

\smallskip

\noindent\textbf{Implementation Details.} Each trial we sampled $\theta_r \sim \mathcal{N}(5,0.5)$, $\theta_h \sim \mathcal{N}(1,0.5)$, and $\hat{\theta}_r^0 \sim \mathcal{U}(0.05, 0.15)$. We set the mean value of $\theta_r$ to be higher than $\hat{\theta}_r^0$ to focus on interesting cases where the human initially believed that the robot cared less about its effort than it actually did. The parameters in (\ref{eq:E1}) were $m=0.5$ kg, $b = 1$ N$\cdot$s/m, $\Delta = 0.5$ s, $x^0 = [-1, 0]^\top$, and $H = 10$ s.

\subsection{Dependent Measures}

We report the total cost (i.e., negative reward) of the HRI task: \textit{Robot Cost} is $-R_r$ from (\ref{eq:E2}) and \textit{Human Cost} is $-R_h$ from (\ref{eq:E2}). We also measure how communicative the robot's actions are: a communicative action is an action that pushes the table away from the goal. \emph{Communication} \% refers to the percent of the robot's total effort  $\|\bm{u_r}\|_1$ that pushes the table in the opposite direction of the goal.

\subsection{Communicating with Trusting Humans}

Figs.~\ref{fig:result} and \ref{fig:learn} summarize our results when interacting with Learn humans that trust the robot. \emph{Nash} and \emph{Optimistic} robots always apply positive forces to push the table towards the goal. By contrast, robots which realize that the human trusts them (\emph{Trust}) apply negative forces at the start of the task, moving the table away from the goal. Although seemingly counter-productive, these actions signal to the human that the robot places a high value on its own effort, and increase the human's estimate $\hat{\theta}_r$ so that the human exerts more effort to guide the table towards the goal. Put another way, pulling the table away from the goal convinces the human to take over.

Inspecting Fig.~\ref{fig:learn}, we find that the communicative actions of the \emph{Trust} robot yield lower robot cost than either \emph{Nash} or \emph{Optimistic} across all tested learning rates $\eta$. The communication percent decreases as the human's learning rate increases, since fewer negative forces are required to convey the robot's reward if the human learns quickly. Interestingly, \emph{Trust} also reduced the human cost as compared to \emph{Nash} and \emph{Optimistic}: modeling the human as a trusting agent here led to improved performance for both the human and robot as compared to conservative and optimistic models.

\smallskip

\noindent \textbf{Summary.} Communicative actions emerged as part of the optimal response when interacting with a trusting human; the \emph{Trust} robot intentionally signaled this human by moving away from the goal, resulting in lower overall costs.

\subsection{Human Model Errors}

\begin{figure}[t]

\vspace{0.5em}

	\begin{center}
		\includegraphics[width=0.95\columnwidth]{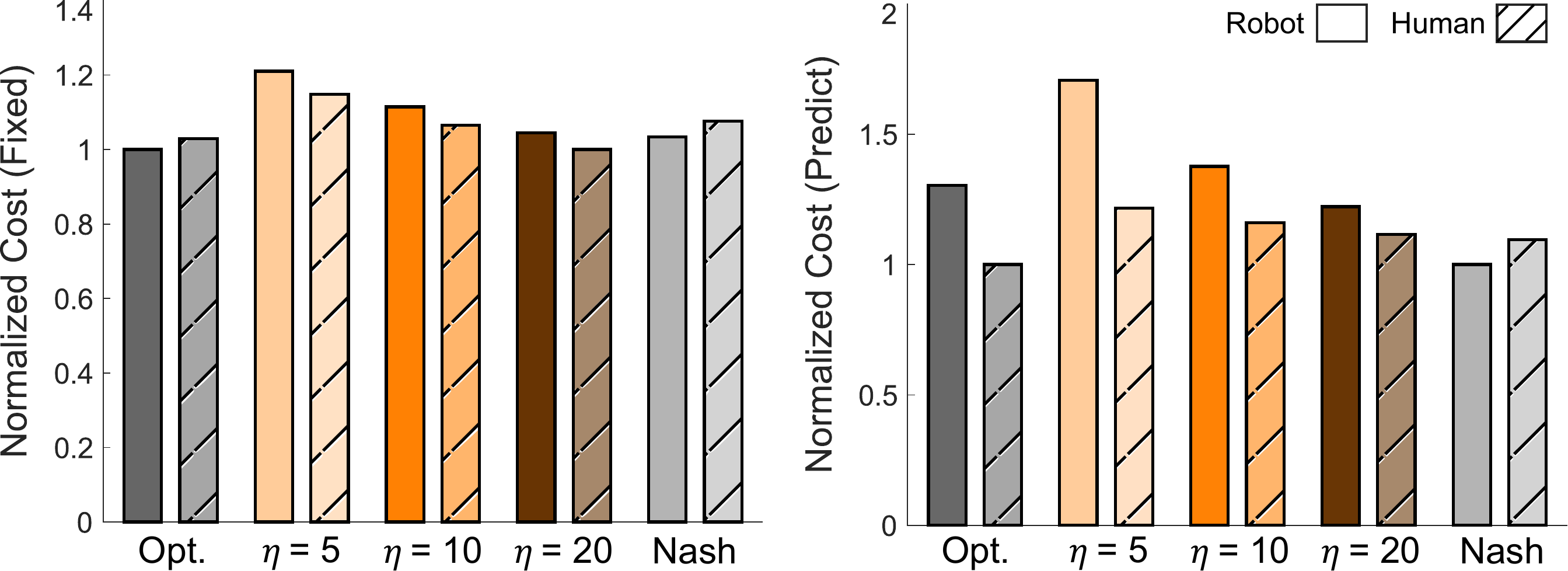}

		\caption{Cost when interacting with non-trusting humans that instead follow our Fixed or Predict models. \emph{Opt} is short for optimistic. The \emph{Trust} robot incorrectly assumes that the human learns with $\eta = \{5, 10, 20\}$. Higher assumed learning rates $\eta$ mitigate errors in the trusting model, since the robot thinks the human learns faster and takes fewer communicative actions.}

		\label{fig:fixed}
	\end{center}

\vspace{-2em}

\end{figure}

So far we have focused on trusting humans that learn from robot actions. In Fig.~\ref{fig:fixed}, however, we simulate users that diverge from this trusting model. If the robot assumes that the human learns---but the human does not---then \emph{Trust} can be outperformed by \emph{Optimistic} and \emph{Nash}. We find that \emph{Trust} becomes more robust to human model errors when the robot assumes higher values of $\eta$, or, put another way, when the robot takes fewer communicative actions. This matches with our intuition: if the robot pushes the table away from the goal to communicate their objective, but the human does not learn from this action, then the robot's communication is inefficient and costly. Hence, when the robot is unsure about the human's trust, the robot should start by assuming a high learning rate $\eta$ to reduce the amount of communication.

\section{User Study}

\begin{figure}[t]

\vspace{0.5em}

	\begin{center}
		\includegraphics[width=0.9\columnwidth]{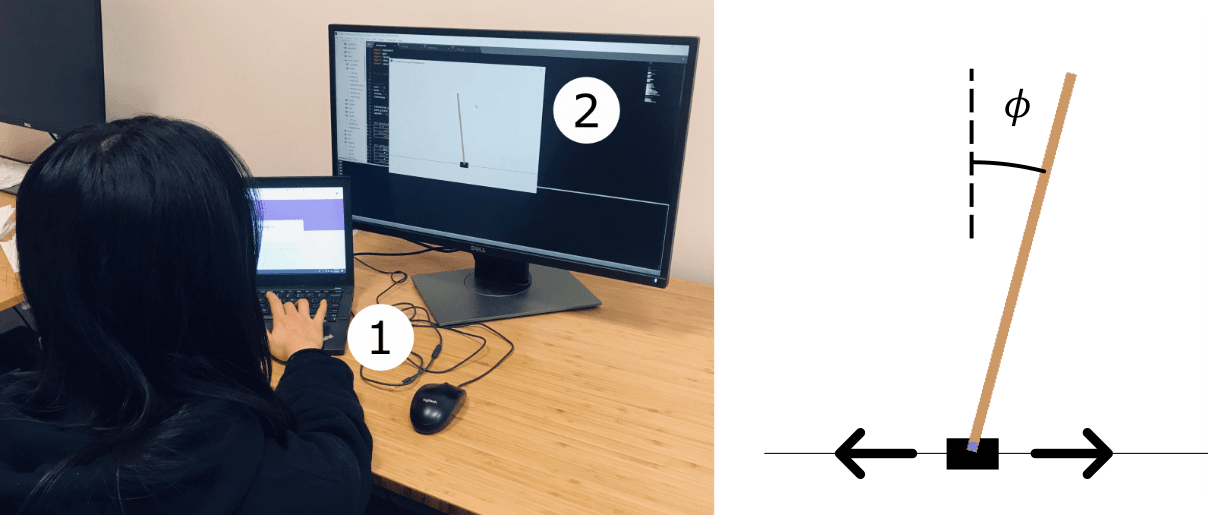}

		\caption{Setup for our user study. Participants used a keyboard (1) to control a inverted pendulum (2, right). A virtual robot also applied forces to control this pendulum; when the human stopped providing inputs, the robot was in charge of keeping the pendulum upright. Robots using our trusting human model intentionally unbalanced the pendulum to increase user involvement.}

		\label{fig:setup}
	\end{center}

\vspace{-0em}

\end{figure}

To test how robots that model uncertain humans as trusting agents interact with \emph{actual} human end-users, we conducted a user study. Our participants worked with a robot to balance a simulated inverted pendulum (see Fig.~\ref{fig:setup}). We compared two different interaction strategies for the robot: computing the Nash equilibrium, where the conservative robot assisted at every timestep to balance the inverted pendulum, and our trusting model, where the robot occasionally unbalanced the pendulum to convince the human to take over. Both the participant and robot wanted to keep the pendulum upright; however, the robot is also rewarded for maximizing end-user involvement, so that the human balances the pendulum. We hypothesized that a robot which reasons over human trust will \emph{increase user involvement} during collaborative tasks.

\smallskip

\noindent \textbf{Independent and Dependent Variables.} We varied the robot interaction strategy with two levels: a conservative \emph{Nash} strategy, which was based on the Nash equilibrium formulation (\ref{eq:fn}), and a communicative \emph{Trust} strategy, which was inspired by our trusting model (\ref{eq:ft}). Let $\phi$ be the angle of the pendulum from vertical; then, $u_r = -sign(\phi)$ for the Nash strategy. Under the Trust strategy, $u_r = sign(\phi)$ at the start of the task to intentionally unbalance the pendulum.

For each interaction strategy we measured the normalized \emph{Time Upright} and \emph{Human Effort}. Time upright is the number of timesteps the pendulum remained vertical, i.e., $|\phi| < \pi/2$. Human effort is defined as the percentage of timesteps that the participant applied inputs to control the cart. Besides these objective measures, we also recorded the participants' subjective responses to a $7$-point Likert scale survey.

\smallskip

\noindent \textbf{Experimental Procedure.} Our participants consisted of $10$ volunteers ($3$ female, $7$ male), with ages ranging from $18-29$ years. Participants used a keyboard to control the 1-DoF cart in the CartPole environment (OpenAI Gym). After a familiarization phase---where the participant interacted without any robot intervention---users completed five trials with the Nash robot strategy and five trials with the Trust robot strategy. We counterbalanced the order of the robot strategies, so that half of the participants interacted with the Trust robot first.

\smallskip

\begin{figure}[t]

	\begin{center}
		\includegraphics[width=0.95\columnwidth]{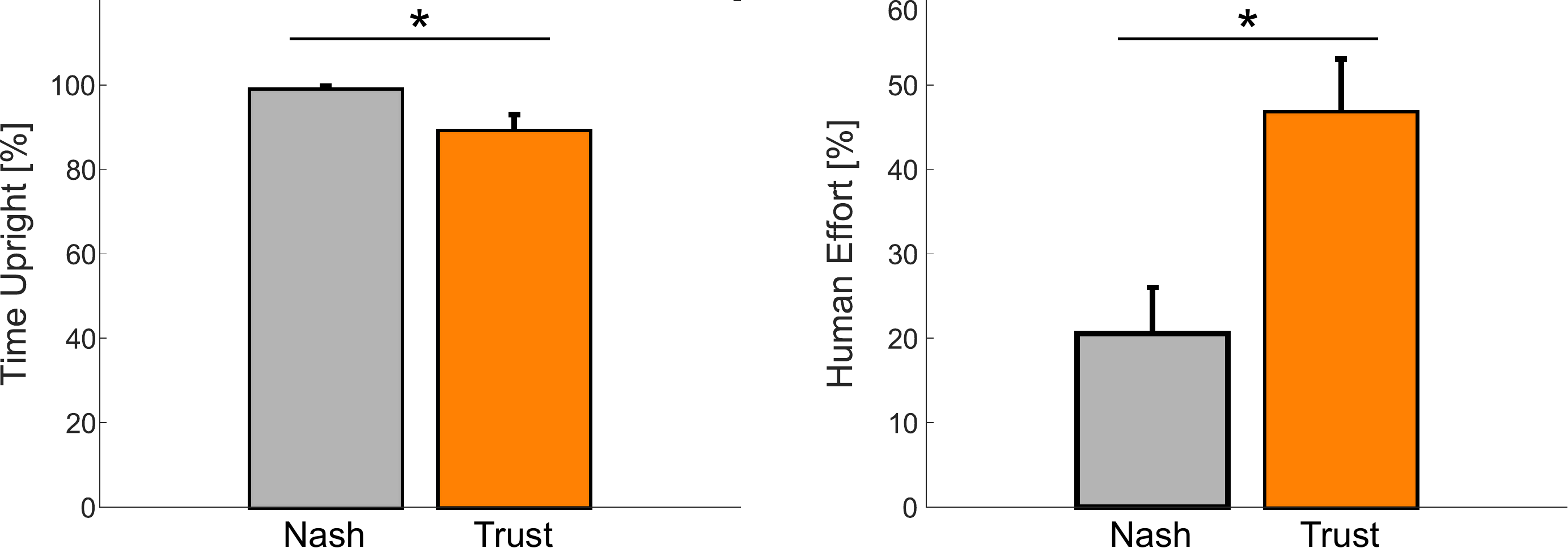}

		\caption{Objective results from our user study. Left: percentage of time the pendulum was upright. Right: percentage of time the human was involved in controlling the pendulum. Error bars show SEM, and an $*$ denotes statistical significance ($p<.05$). The task was more challenging when the robot modeled the human as trusting (as evidenced by the lower time upright), but the user became more involved, and interacted for a longer duration.}

		\label{fig:user}
	\end{center}

\vspace{-2em}

\end{figure}

\noindent \textbf{Results.} The results of our user study are summarized in Fig.~\ref{fig:user} and Table~\ref{table2}. To interpret our objective results we performed paired t-tests with robot strategy as the within-subjects factor. We found a statistically significant difference between Nash and Trust for time upright ($t(9)=3.1$, $p < .05$), as well as for human effort ($t(9)=4.8$, $p < .001$). The communicative actions taken by the Trust robot (intentionally unbalancing the pendulum) led to greater human involvement.

We grouped our survey questions into three topics: understanding, objectives, and adaptation. Paired t-tests revealed that there was not a statistically significant difference in how well participants understood Nash and Trust ($t(19)=0.9$, $p = .38$). Despite this confidence, users incorrectly believed that Trust did not want to keep the pendulum upright ($t(19)=4.5$, $p = <.01$). Users also reported that they adapted their approach more when interacting with Trust ($t(19)=3.0$, $p = <.01$). Taken together, these responses suggest that Trust successfully mislead users: by unbalancing the pendulum, the robot convinced participants that it had a different objective, and caused users to adjust their behavior.

\begin{table}[t]

\vspace{1em}
	\caption{Results of our $7$-point Likert scale survey. The questions were grouped by understanding, objectives, and adaptation. Higher values indicate agreement with the positive statement.}
	\label{table2}
	\begin{center}
		\begin{tabular}{{l}{c}{c}{c}}
\textbf{Questions} & \textbf{Nash} & \textbf{Trust} & \textbf{p-value} \bigstrut \\ \hline
$+$ I could tell what the robot & \multirow{4}{*}{5.45} & \multirow{4}{*}{5.05} & \multirow{4}{*}{0.38} \bigstrut[t] \\
wanted to achieve & & & \\
$-$ I did not \emph{understand} the & & & \\
robot's objective & & & \bigstrut[b] \\ \hline
$+$ The robot's goal was to keep & \multirow{4}{*}{5.60} & \multirow{4}{*}{3.45} & \multirow{4}{*}{$<.001^*$} \bigstrut[t] \\
the pendulum upright & & & \\
$-$ The robot had a different & & & \\
\emph{objective} than I did & & & \bigstrut[b] \\ \hline
$+$ I \emph{adapted} over time to & \multirow{4}{*}{3.75} & \multirow{4}{*}{5.1} & \multirow{4}{*}{$<.01^*$} \bigstrut[t] \\
the robot's actions & & & \\
$-$ I used the same strategy  & & & \\
across trials & & & 
		\end{tabular}
	\end{center}

	\vspace{-2.5em}	

\end{table}

\section{Discussion}

\noindent \textbf{Summary.} We focused on partial information settings where the robot knows both the human's objective and its own, while the human only knows their own objective. The robot's behavior within these settings depends on how it models the human. We formulated four different models for two-player HRI games: the robot can behave conservatively (as if the human knew their objective), optimistically (as if they can teach the human any objective), model the human as rational, or treat the human as a trusting agent. This trusting model is a simplification of the rational model, where the human learns while assuming that the robot behaves conservatively.

Comparing our trusting human model to the alternatives, we found that modeling the human as trusting is not the same as acting optimistically, but can result in robots that take advantage of the human's trust. In a linear-quadratic case study robots with a trusting human model intentionally performed locally sub-optimal actions (moving away from the goal) to convince the human to take over. During our user study, robots with the trusting human model similarly convinced humans that the robot did not share their objective, resulting in adaptation and increased user involvement.

\smallskip

\noindent \textbf{Why Trusting Models?} Programming robots to model humans as trusting agents has benefits as a way to generate communicative actions and increase user involvement. Under the trusting model, communicative actions naturally emerge as part of the solution to an MDP, and do not need to be explicitly encoded or rewarded. Our case and user studies suggest that these actions increase user involvement, which can be particularly useful for assistive robotics and rehabilitation applications, where \emph{over-assistance} is a common problem \cite{blank2014}. We think robots that are aware of the human's trust can make intelligent decisions to avoid behaviors such as over-assistance.  

\smallskip

\noindent \textbf{Limitations and Future Work.} Our work is a first step towards developing influential and communicative robot behavior in settings where the human is uncertain. We have so far assumed that the human and robot coordinate to follow the same Nash equilibrium; in general, this is not the case, and so complete solutions must also account for strategy uncertainty.
We also recognize that computing the Nash equilibria to these two-player games with high-dimensional states and actions can quickly become computationally infeasible. 

These limitations suggest that---in order to implement our trust model in realistic settings---we need to identify suitable, real-time approximations. Our future work focuses on this challenge: we plan to bias the human model towards \emph{trusting but myopic} human behavior, which we can obtain tractably without identifying Nash equilibria. Equipped with this approximation, we will pursue realistic applications such as shared control and assistive robotics.

\section{Acknowledgements}

Toyota Research Institute (``TRI")  provided funds to assist the authors with their research but this article solely reflects the opinions and conclusions of its authors and not TRI or any other Toyota entity. The authors would also like to acknowledge General Electric (GE) for their support.

\bibliographystyle{IEEEtran}
\bibliography{IEEEabrv,citations}

\begin{thebibliography}{10}
\providecommand{\url}[1]{#1}
\csname url@samestyle\endcsname
\providecommand{\newblock}{\relax}
\providecommand{\bibinfo}[2]{#2}
\providecommand{\BIBentrySTDinterwordspacing}{\spaceskip=0pt\relax}
\providecommand{\BIBentryALTinterwordstretchfactor}{4}
\providecommand{\BIBentryALTinterwordspacing}{\spaceskip=\fontdimen2\font plus
\BIBentryALTinterwordstretchfactor\fontdimen3\font minus
  \fontdimen4\font\relax}
\providecommand{\BIBforeignlanguage}[2]{{%
\expandafter\ifx\csname l@#1\endcsname\relax
\typeout{** WARNING: IEEEtran.bst: No hyphenation pattern has been}%
\typeout{** loaded for the language `#1'. Using the pattern for}%
\typeout{** the default language instead.}%
\else
\language=\csname l@#1\endcsname
\fi
#2}}
\providecommand{\BIBdecl}{\relax}
\BIBdecl

\bibitem{baker2009}
C.~L. Baker, R.~Saxe, and J.~B. Tenenbaum, ``Action understanding as inverse
  planning,'' \emph{Cognition}, vol. 113, no.~3, pp. 329--349, 2009.

\bibitem{blank2014}
A.~A. Blank, J.~A. French, A.~U. Pehlivan, and M.~K. O’Malley, ``Current
  trends in robot-assisted upper-limb stroke rehabilitation: {P}romoting
  patient engagement in therapy,'' \emph{Current Physical Medicine and
  Rehabilitation Reports}, vol.~2, no.~3, pp. 184--195, 2014.

\bibitem{hancock2011}
P.~A. Hancock, D.~R. Billings, K.~E. Schaefer, J.~Y. Chen, E.~J. De~Visser, and
  R.~Parasuraman, ``A meta-analysis of factors affecting trust in human-robot
  interaction,'' \emph{Human Factors}, vol.~53, no.~5, pp. 517--527, 2011.

\bibitem{xu2015}
A.~Xu and G.~Dudek, ``Optimo: {O}nline probabilistic trust inference model for
  asymmetric human-robot collaborations,'' in \emph{ACM/IEEE Int. Conf.
  Human-Robot Interaction (HRI)}, 2015, pp. 221--228.

\bibitem{wang2016}
N.~Wang, D.~V. Pynadath, and S.~G. Hill, ``Trust calibration within a
  human-robot team: {C}omparing automatically generated explanations,'' in
  \emph{Int. Conf. Human Robot Interaction (HRI)}, 2016, pp. 109--116.

\bibitem{chen2018}
M.~Chen, S.~Nikolaidis, H.~Soh, D.~Hsu, and S.~Srinivasa, ``Planning with trust
  for human-robot collaboration,'' in \emph{ACM/IEEE Int. Conf. on Human-Robot
  Interaction (HRI)}, 2018, pp. 307--315.

\bibitem{hellstrom2018}
T.~Hellstr{\"o}m and S.~Bensch, ``Understandable robots - what, why, and how,''
  \emph{Paladyn, Journal of Behavioral Robotics}, vol.~9, no.~1, pp. 110--123,
  2018.

\bibitem{sadigh2016}
D.~Sadigh, S.~Sastry, S.~A. Seshia, and A.~D. Dragan, ``Planning for autonomous
  cars that leverage effects on human actions.'' in \emph{Robotics: Science and
  Systems (RSS)}, 2016.

\bibitem{huang2017}
S.~H. Huang, D.~Held, P.~Abbeel, and A.~D. Dragan, ``Enabling robots to
  communicate their objectives,'' \emph{Autonomous Robots}, pp. 1--18, 2017.

\bibitem{losey2019}
D.~Losey and M.~K. O'Malley, ``Enabling robots to infer how end-users teach and
  learn through human-robot interaction,'' \emph{IEEE Robotics and Automation
  Letters}, vol.~4, no.~2, pp. 1956--1963, 2019.

\bibitem{sadigh2018planning}
D.~Sadigh, N.~Landolfi, S.~S. Sastry, S.~A. Seshia, and A.~D. Dragan,
  ``Planning for cars that coordinate with people: Leveraging effects on human
  actions for planning and active information gathering over human internal
  state,'' \emph{Autonomous Robots (AURO)}, vol.~42, no.~7, pp. 1405--1426,
  October 2018.

\bibitem{dragan2014}
A.~D. Dragan, R.~M. Holladay, and S.~S. Srinivasa, ``An analysis of deceptive
  robot motion.'' in \emph{RSS}, 2014, p.~10.

\bibitem{foerster2018}
J.~Foerster, R.~Y. Chen, M.~Al-Shedivat, S.~Whiteson, P.~Abbeel, and
  I.~Mordatch, ``Learning with opponent-learning awareness,'' in \emph{AAMAS},
  2018, pp. 122--130.

\bibitem{berlin2006}
M.~Berlin, J.~Gray, A.~L. Thomaz, and C.~Breazeal, ``Perspective taking: {A}n
  organizing principle for learning in human-robot interaction,'' in
  \emph{AAAI}, 2006, pp. 1444--1450.

\bibitem{pandey2013}
A.~K. Pandey, M.~Ali, and R.~Alami, ``Towards a task-aware proactive sociable
  robot based on multi-state perspective-taking,'' \emph{International Journal
  of Social Robotics}, vol.~5, no.~2, pp. 215--236, 2013.

\bibitem{trafton2005}
J.~G. Trafton, N.~L. Cassimatis, M.~D. Bugajska, D.~P. Brock, F.~E. Mintz, and
  A.~C. Schultz, ``Enabling effective human-robot interaction using
  perspective-taking in robots,'' \emph{IEEE Transactions on Systems, Man, and
  Cybernetics-Part A: Systems and Humans}, vol.~35, no.~4, pp. 460--470, 2005.

\bibitem{osborne1994}
M.~J. Osborne and A.~Rubinstein, \emph{A Course in Game Theory}.\hskip 1em plus
  0.5em minus 0.4em\relax MIT Press, 1994.

\bibitem{hedden2002}
T.~Hedden and J.~Zhang, ``What do you think {I} think you think?: {S}trategic
  reasoning in matrix games,'' \emph{Cognition}, vol.~85, no.~1, pp. 1--36,
  2002.

\bibitem{simon1997}
H.~A. Simon, \emph{Models of bounded rationality: {E}mpirically grounded
  economic reason}.\hskip 1em plus 0.5em minus 0.4em\relax MIT press, 1997,
  vol.~3.

\bibitem{puterman2014}
M.~L. Puterman, \emph{Markov Decision Processes: {D}iscrete Stochastic Dynamic
  Programming}.\hskip 1em plus 0.5em minus 0.4em\relax John Wiley \& Sons,
  2014.

\bibitem{liu2017}
W.~Liu, B.~Dai, A.~Humayun, C.~Tay, C.~Yu, L.~B. Smith, J.~M. Rehg, and
  L.~Song, ``Iterative machine teaching,'' in \emph{Int. Conf. on Machine
  Learning (ICML)}, 2017.

\bibitem{engwerda2007}
J.~Engwerda, ``Algorithms for computing {N}ash equilibria in deterministic {LQ}
  games,'' \emph{Computational Management Science}, vol.~4, no.~2, pp.
  113--140, 2007.

\bibitem{li2019}
Y.~Li, G.~Carboni, F.~Gonzalez, D.~Campolo, and E.~Burdet, ``Differential game
  theory for versatile physical human--robot interaction,'' \emph{Nature
  Machine Intelligence}, vol.~1, no.~1, p.~36, 2019.

\end{thebibliography}

\end{document}